\documentclass[11pt]{article}

\usepackage[preprint]{acl}

\usepackage{times}
\usepackage{latexsym}

\usepackage[T1]{fontenc}
\usepackage{amsmath, amssymb}
\usepackage{booktabs}
\usepackage{subcaption}

\usepackage[utf8]{inputenc}

\usepackage{microtype}

\usepackage{inconsolata}

\usepackage{graphicx}

\title{Soft Sequence Policy Optimization}

\author{
Svetlana Glazyrina$^{1}$ \,
Maksim Kryzhanovskiy$^{1,2}$ \,
Roman Ischenko$^{1,2}$ \\
\\
$^{1}$ Lomonosov Moscow State University, Moscow, Russia \\
$^{2}$ Institute for Artificial Intelligence, Lomonosov Moscow State University, Moscow, Russia
}

\begin{document}
\maketitle
\begin{abstract}
A significant portion of recent research on Large Language Model (LLM) alignment focuses on developing new policy optimization methods based on Group Relative Policy Optimization (GRPO). Two prominent directions have emerged: (i) a shift toward sequence-level importance sampling weights that better align with the sequence-level rewards used in many tasks, and (ii) alternatives to the PPO-style clipping that aim to avoid the associated loss of training signal and entropy collapse. We introduce Soft Sequence Policy Optimization, an off-policy reinforcement learning objective that incorporates soft gating functions over token-level probability ratios within sequence-level importance weights. We provide theoretical motivation for SSPO and investigate practical modifications to improve optimization behavior. Empirically, we demonstrate that SSPO improves training stability and performance both in mathematical reasoning and coding tasks.
\end{abstract}

\section{Introduction}

Reinforcement learning (RL) has become a central ingredient in enhancing the reasoning abilities of large language models (LLMs), particularly for tasks requiring long chains of thought (CoT), multi-step decision making, and delayed rewards. While supervised fine-tuning excels at imitating local patterns present in static datasets, RL enables optimization over entire generated sequences, making it especially attractive for complex reasoning, program synthesis, and decision-centric language modeling. 

Among modern RL approaches for LLM alignment, group-based policy optimization methods have emerged as a practical and effective strategy. These methods sample multiple candidate responses per input prompt and normalize sequence-level rewards within each group, yielding a reliable prompt value estimate without relying on an auxiliary critic network. Representative algorithms such as Group Relative Policy Optimization (GRPO) \citep{shao2024grpo} and REINFORCE Leave-One-Out (RLOO) \citep{ahmadian2024rloo} demonstrate that relative comparisons within a group can significantly stabilize training, reduce computational overhead, and improve reasoning performance across a wide range of domains. 

Despite their empirical success, existing group-based methods face important limitations when deployed at scale. In particular, off-policy learning becomes unavoidable in realistic training pipelines. As model sizes increase, architectures incorporate sparsity (e.g., Mixture-of-Experts), and generated sequences become longer, large rollout batches are required to fully utilize modern hardware. To improve sample efficiency, these batches are typically partitioned into multiple mini-batches for gradient updates. This practice inevitably induces an off-policy setting, where responses are sampled from a behavior policy while updates are applied to a newer policy, a regime in which GRPO-like algorithms increasingly dominate current practice.

In off-policy group-based optimization, policy updates are usually weighted by importance sampling (IS) ratios between the current and behavior policies. These ratios are well known to suffer from high variance, especially for long sequences where token-level likelihood ratios compound multiplicatively. Uncontrolled IS weights can destabilize training and ultimately degrade performance. Two broad strategies have emerged to mitigate this issue.

First, many methods apply hard clipping to large importance weights, limiting their influence on the gradient \citep{schulman2017ppo, shao2024grpo, minimax2025cispo}. While effective at reducing variance, hard clipping introduces a difficult tradeoff: aggressive clipping improves stability but reduces sample efficiency and limits exploration \citep{chen2022sufficiencyoffpolicynesssoftclipping, dwyer2025itsyouitsclipping}, whereas loose clipping preserves learning signal at the cost of noisy and brittle updates. 

Second, in accordance with insights from RLOO and motivated by the prevalence of sequence-level rewards in LLM alignment (e.g., RLHF and RLVR), recent work proposes enforcing the sequence-level coherence of importance weights rather than treating token-level contributions independently. 
This perspective motivates objectives such as Group Sequence Policy Optimization (GSPO) \citep{zheng2025gspo} and Geometric-Mean Policy Optimization (GMPO) \citep{zhao2025gmpo}, which replace arithmetic aggregation of token-level ratios with multiplicative or geometric formulations that better respect the structure of sequence probabilities. A key innovation of GMPO is the combination of sequence-level weighting with token-level probability ratio clipping.

However, existing solutions remain incomplete. Sequence-coherent objectives, such as GSPO, improve stability but do not fully address the interaction between off-policy learning and entropy-regularized objectives commonly used in modern RL for LLMs. In parallel, soft policy optimization methods such as Soft Adaptive Policy Optimization (SAPO) \citep{gao2025sapo} highlight the importance of entropy-aware objectives that smoothly interpolate between exploitation and exploration. However, they are not explicitly designed for sequence-level optimization.

In this paper, we propose \emph{Soft Sequence Policy Optimization (SSPO)}, a new off-policy reinforcement learning objective that unifies and extends insights from reward coherent sequence weights and soft policy optimization. SSPO introduces a soft sequence-level weighting mechanism with token-level weight attenuation that controls importance sampling variance without resorting to hard clipping while preserving coherent credit assignment to entire responses. By operating at the sequence level and incorporating entropy-aware regularization, SSPO achieves a more favorable exploration-exploitation tradeoff than prior approaches in off-policy group-based RL.

Our contributions are as follows:
\begin{itemize}
\item We analyze the properties of token-wise gate functions when importance weights are aggregated via the geometric mean.
\item We propose Soft Sequence Policy Optimization (SSPO), a sequence-level objective with soft clipping for off-policy reinforcement learning.
\item We demonstrate improved training stability and performance on various model sizes, architectures, and domains.
\end{itemize}

\section{Preliminaries}

Since we do not modify the additive KL-divergence regularization term in the loss, we omit it for brevity.

\paragraph{Notation}
Following standard conventions in the literature, an autoregressive language model parameterized by $\theta$ is defined by a policy $\pi_\theta$. $x$ denotes a query and $\mathcal{D}$ denotes the set of queries. Given a response $y$ to a query $x$, the likelihood of $y$ under the policy $\pi_\theta$ is defined as
$$
\pi_\theta (y \mid x) = \prod_{t=1}^{|y|} \pi_\theta (y_t \mid x, y_{<t}),
$$
where $|y|$ denotes the number of tokens in $y$. Each query-response pair $(x, y)$ is assigned a reward $r(x, y)$ by a verifier $r$. 

Throughout this paper, we adopt a compact notation for advantage-dependent clipping operators commonly used in policy optimization objectives. 
Rather than explicitly writing expressions of the form
$$
\min \!\Big(
\rho \cdot\widehat{A},\;
\mathrm{clip}\!\left(\rho,\,1-\varepsilon_{\mathrm{low}},\,1+\varepsilon_{\mathrm{high}}\right)\cdot\widehat{A}
\Big),
$$
we define the advantage-aware clipping function
\begin{equation}
f_{\mathrm{Clip}}(\rho; \widehat{A})
\triangleq
\begin{cases}
\min(\rho,1+\varepsilon_{\mathrm{high}}), & \widehat{A} > 0,\\
\max(\rho,1-\varepsilon_{\mathrm{low}}), & \widehat{A} \le 0,
\end{cases}
\label{eq:clip_def}
\end{equation}
and write the clipped weights compactly as
$f_{\mathrm{Clip}}(\rho;\,\widehat{A})\,\cdot \widehat{A}$.

This notation enables a unified presentation of hard and soft clipping mechanisms in the subsequent sections.

\paragraph{Group Relative Policy Optimization}
Building on Proximal Policy Optimization (PPO) \citep{schulman2017ppo}, Group Relative Policy Optimization (GRPO) \citep{shao2024grpo} enables off-policy learning by leveraging samples generated by an older (behavior) policy $\pi_{\theta_{\mathrm{old}}}$. Following PPO, GRPO employs a clipping mechanism to constrain policy updates within a proximal region around the behavior policy. The key difference is that GRPO replaces computationally and memory-intensive advantage estimators based on a critic network, whose reliability may be questionable \citep{vineppo}, with a relative advantage computed among responses to the same query. The GRPO optimization objective is
\begin{multline}
\label{equ:grpo}
\mathcal{J}_\text{GRPO}(\theta) = \mathbb{E}_{ x \sim \mathcal{D},\, \{y_i\}_{i=1}^G \sim \pi_{\theta_\text{old}}( \cdot | x) } \\
\left[ \frac{1}{G} \sum_{i=1}^{G} \frac{1}{|y_i|} \sum_{t=1}^{|y_i|} 
f_{\mathrm{Clip}}(\rho_{i, t}(\theta); \, \widehat{A}_{i, t})\cdot \widehat{A}_{i, t}
\right].
\end{multline}
Here, $G$ denotes the group size, i.e., the number of responses generated for a single query.  The importance ratio $\rho_{i,t}(\theta)$ and the advantage $\widehat{A}_{i,t}$ for token $y_{i,t}$ are defined as
\begin{gather}
    \rho_{i,t}(\theta)=\frac{ \pi_{\theta} (y_{i,t} | x, y_{i,<t}) }{ \pi_{\theta_\text{old}} (y_{i,t} | x,y_{i,<t})},\\
    \widehat{A}_{i,t} = \widehat{A}_{i} = \frac{r(x, y_i) - \mathrm{mean} \left( \{ r(x, y_i) \}_{i=1}^G \right) }{ \mathrm{std} \left( \{ r(x, y_i) \}_{i=1}^G \right) }.
\end{gather}

The notation $\widehat{A}_{i} = \widehat{A}_{i,t}$ is valid since all tokens in the sequence $y_i$ share the same advantage.

\paragraph{Group Sequence Policy Optimization}
Group Sequence Policy Optimization (GSPO) \citep{zheng2025gspo} identifies a primary problem of GRPO: a mismatch between the unit of optimization and the unit of reward. In GRPO, importance sampling weights for off-policy correction and clipping are applied at the token level, whereas the advantage remains constant across all tokens in a sequence. In contrast, GSPO performs optimization, and in particular clipping, directly at the sequence level. Its objective is given by
\begin{multline}
\mathcal{J}_{\text{GSPO}}(\theta) =
\mathbb{E}_{x \sim \mathcal{D},\, \{y_i\}_{i=1}^G \sim \pi_{\theta_{\mathrm{old}}}(\cdot \mid x)} \\
\left[
\frac{1}{G} \sum_{i=1}^{G}
f_{\mathrm{Clip}}(s_i(\theta); \, \widehat{A}_i)\cdot \widehat{A}_i
\right],
\label{equ:gspo}
\end{multline}
where the importance ratio $s_i(\theta)$ is computed based on the sequence likelihood, with length normalization to control variance \citep{zheng2023click}:
\begin{align}
s_{i}(\theta) = \left( \frac{ \pi_{\theta} (y_i | x) }{ \pi_{\theta_\text{old}} (y_i | x)} \right)^{\frac{1}{|y_i|}}
\end{align}

GSPO has been shown to improve training stability for large models and to increase sample efficiency compared to GRPO, although the fraction of clipped tokens grows.

\paragraph{Geometric-Mean Policy Optimization}

Geometric-Mean Policy Optimization (GMPO) \citep{zhao2025gmpo} combines response-level importance sampling with token-level probability ratio clipping. GMPO adopts the same sequence-length normalization as GSPO but is motivated by a different intuition: reducing variance in token-level importance sampling weights and sensitivity to outliers by replacing the arithmetic mean over tokens in the GRPO objective with a geometric mean. Specifically, GMPO optimizes the following objective:
\begin{multline}
\mathcal{J}_{\mathrm{GMPO}}(\theta) = \mathbb{E}_{x\sim \mathcal{D}, \{y_i\}_{i=1}^G \sim \pi_{\theta_{\mathrm{old}}}(\cdot|x)} \\ 
 \left[ \frac{1}{G}\sum_{i=1}^{G}\left[\prod_{t=1}^{|y_i|}
 \Big| f_{\mathrm{Clip}}(\rho_{i, t}(\theta);  \widehat{A}_i)
 \Big|
\right]^{\frac{1}{|y_i|}}\cdot \hat{A}_{i} \right].
\label{eq:gmpo}
\end{multline}

This geometric aggregation improves robustness to tokens with disproportionately large importance weights and  reduces the range of the loss values, leading to more stable updates. The authors further observe that token-level clipping is less aggressive than sequence-level clipping and more effectively constrains the range of importance sampling ratios.

\paragraph{Soft Adaptive Policy Optimization}

In existing group-based policy optimization methods, high variance in importance sampling weights is typically mitigated through hard clipping. This introduces a trade-off between update stability and sample efficiency and may also reduce exploration abilities. As an alternative, Soft Adaptive Policy Optimization (SAPO) \citep{gao2025sapo} employs a smooth, temperature-controlled gating mechanism that preserves the learning signal for all sampled tokens. The SAPO objective is defined as:
\begin{multline}
\mathcal{J}_{\mathrm{SAPO}}(\theta) = \mathbb{E}_{x\sim \mathcal{D}, \{y_i\}_{i=1}^G\sim \pi_{\theta_\text{old}}(\cdot\mid x)} \\
\Bigg[\frac{1}{G} \sum_{i=1}^{G} \frac{1}{|y_i|} \sum_{t=1}^{|y_i|} 
 f_{\mathrm{Soft}}(\rho_{i,t}(\theta) ;\, \widehat{A}_{i, t})
\cdot \widehat{A}_{i,t}
 \Bigg],
\label{eq:sapo}    
\end{multline}

where
    \begin{gather}
f_\mathrm{Soft}(x; \widehat{A}) = \sigma\left(\tau(\widehat{A})\,\cdot (x-1)\right) \cdot \frac{4}{\tau(\widehat{A})}, \; \\
\tau(\widehat{A}) = 
    \begin{cases} 
    \tau_{\text{pos}}, & \text{if } \widehat{A} > 0\\
    \tau_{\text{neg}}, & \text{otherwise} 
    \end{cases}.
\label{eq:sapo_gate}
\end{gather}

Here, $\tau_{\mathrm{pos}}$ and $\tau_{\mathrm{neg}}$ denote the temperatures for positive and negative advantages, respectively, and $\sigma(x)=1/(1+e^{-x})$ is the sigmoid function. 

By construction, the sigmoid-shaped gating function preserves gradients during on-policy updates while remaining bounded. Attenuating the contribution of outlier tokens, rather than truncating it entirely, allows SAPO to maintain informative learning signals throughout training and induces a continuous trust region.

\section{Soft Sequence Policy Optimization}

While Soft Adaptive Policy Optimization (SAPO) preserves token-level adaptivity through smooth gating, it is sequence-coherent only under two mild assumptions:
(i) small policy updates, i.e., $\rho_{i,t}(\theta) \approx 1$, and 
(ii) low intra-sequence dispersion of importance ratios,
$
\frac{1}{|y_i|}\sum_{t=1}^{|y_i|} \bigl(\log \rho_{i,t}(\theta) - \log s_i(\theta)\bigr)^2$.
These assumptions may be violated in practice, particularly for long sequences or during the early stages of training. We therefore aim to redesign the SAPO objective to relax these assumptions while retaining token-level adaptivity.

In the absence of clipping, the geometric mean of token-wise importance ratios coincides with the length-normalized sequence-level importance ratio used in GSPO. GMPO can therefore be interpreted as a relaxation of sequence-level policy optimization that preserves token-level adaptivity.

Building on this observation, we propose \emph{Soft Sequence Policy Optimization} (SSPO). SSPO aggregates token-level gating functions using a geometric mean, yielding the objective
\begin{multline}
\mathcal{J}_{\mathrm{SSPO}}(\theta) = \mathbb{E}_{x\sim \mathcal{D},\, \{y_i\}_{i=1}^G \sim \pi_{\theta_{\mathrm{old}}}(\cdot \mid x)} \\
\left[ \frac{1}{G} \sum_{i=1}^{G} \left( \prod_{t=1}^{|y_i|}   f\!\left(\rho_{i,t}(\theta); \, \widehat{A}_i\right) \right)^{\frac{1}{|y_i|}} \cdot \widehat{A}_{i} \right].
\label{eq:sspo}    
\end{multline}

Here, $f(\, \cdot \,; \widehat{A})$ is a non-negative gating function whose properties are specified below.

\paragraph{Gradient Analysis}
To analyze the optimization behavior of SSPO, we compare its gradient to that of SAPO.

Differentiating the SAPO objective in Eq.~\eqref{eq:sapo} with respect to the model parameters $\theta$ yields the weighted policy gradient
\begin{multline}
\nabla_{\theta}\mathcal{J}_{\mathrm{SAPO}}(\theta)
=\mathbb{E}_{x\sim \mathcal{D},\, \{y_i\}_{i=1}^G\sim \pi_{\theta_{\mathrm{old}}}(\cdot\mid x)} \\
\Bigg[
\frac{1}{G}
\sum_{i=1}^{G}
\frac{1}{|y_i|}
\sum_{t=1}^{|y_i|}
f'_{\mathrm{Soft}}\!\left(\rho_{i,t}(\theta);\widehat{A}_{i,t}\right)
\,\rho_{i,t}(\theta)\, \cdot \\
\cdot \nabla_\theta
\log \pi_\theta(y_{i,t}\mid x,y_{i,<t})
\,\widehat{A}_{i,t}
\Bigg].
\label{eq:sapo_gradient}
\end{multline}
For the sigmoid-shaped gate in SAPO,
\begin{equation}
\begin{gathered}
f'_{\mathrm{Soft}}(\rho ;\widehat{A}) = 4\,g(\rho;\widehat{A})\bigl(1-g(\rho;\widehat{A})\bigr),\\
g(\rho;\widehat{A}) = \sigma\!\left(\tau(\widehat{A})(\rho-1)\right),
\end{gathered}
\label{eq:sapo_weight}
\end{equation}
the derivative peaks at $\rho=1$ and exponentially decays as $\rho$ deviates from unity, inducing a soft trust region while preserving on-policy behavior.

Differentiating the SSPO objective in Eq.~\eqref{eq:sspo} yields
\begin{multline}
\nabla_\theta \mathcal{J}_{\mathrm{SSPO}}(\theta) =
\mathbb{E}_{x\sim \mathcal{D}, \, \{y_i\}_{i=1}^G\sim \pi_{\theta_\mathrm{old}}(\cdot\mid x)} \\
\Bigg[
\frac{1}{G}
\sum_{i=1}^{G}
\left(
\prod_{t=1}^{|y_i|}
f(\rho_{i,t}(\theta);\widehat{A}_i)
\right)^{\!\frac{1}{|y_i|}} \cdot \widehat{A}_i \;\cdot\\
\cdot
\frac{1}{|y_i|}
\sum_{t=1}^{|y_i|}
\frac{f'(\rho_{i,t}(\theta);\widehat{A}_i)}{f(\rho_{i,t}(\theta);\widehat{A}_i)}
\,\rho_{i,t}(\theta)\, \\
\nabla_\theta 
\log \pi_\theta(y_{i,t}\mid x,y_{i,<t})
\Bigg].
\label{eq:sspo_grad}
\end{multline}

Thus, each token-level policy gradient is modulated by two factors: a sequence-level geometric aggregation of token gates and a local soft importance weight defined as
$w(\rho;\widehat{A}) \triangleq \frac{f'(\rho;\widehat{A})}{f(\rho;\widehat{A})}\,\rho$.

\paragraph{Gate Design}
Motivated by Eq.~\eqref{eq:sspo_grad} and the Scopic objective \citep{chen2022sufficiencyoffpolicynesssoftclipping}, we require the gating function family $f(\rho;\widehat{A}): \mathbb{R}_{++} \to \mathbb{R}_{+}$ to satisfy the following conditions:
(i) $f$ is continuously differentiable;
(ii) $f(1;\widehat{A}) = 1$ and $f'(1;\widehat{A}) = 1$;
(iii) the local weight multiplier forms a bell-shaped curve centered at $\rho=1$ to suppress outlier importance ratios;
(iv) $f$ is monotonically non-decreasing.

Condition (ii) ensures that when $\rho_{i,t}(\theta) = 1$, the gradient contribution coincides with the standard on-policy update. Conditions (ii) and (iv) ensure that the gating function preserves the ordinal relationship of the importance ratios, maintaining consistency with sequence-level importance weighting. Condition (iii) enforces controlled attenuation for samples deviating from $\rho=1$, stabilizing training by discouraging aggressive updates from off-policy samples. Furthermore, requiring $w(\rho;\widehat{A}) \to 0$ as $\rho \to \infty$ guarantees that samples with extreme importance ratios contribute negligibly to the gradient, effectively mitigating instability caused by heavy-tailed ratio distributions.

We explore two formalizations of condition (iii). The first (iii.1) requires the weight multiplier $\frac{f'}{f}$ to attain a global maximum at $\rho = 1$ and decrease monotonically as $\rho$ deviates from 1, with $\frac{f'}{f} \to 0$ as $\rho \to \infty$. The second (iii.2) imposes the same properties on the full token-wise weight $w(\rho;\widehat{A}) = \frac{f'(\rho;\widehat{A})}{f(\rho;\widehat{A})}\cdot \rho$, adding $w(0;\widehat{A}) = 0$. We discuss the trade-offs between these formulations in Subsection \ref{sec:gate_choice}.

Based on empirical results provided further, we instantiate the gate as
\begin{equation}
f_{\mathrm{SSPO}}(\rho;\widehat{A})
=
\exp\left(\tau(\widehat{A}) \cdot \arctan\left(\frac{\log \rho}{\tau(\widehat{A})}\right)\right),
\label{eq:sspo_gate}
\end{equation}
where $\tau(\widehat{A})$ is an advantage-dependent temperature. This choice yields a local weight:
\begin{equation}
w(\rho;\widehat{A}) =
\frac{1}{1 + \left(\frac{\log \rho}{\tau(\widehat{A})}\right)^2},
\label{eq:sspo_weight}
\end{equation}
which peaks at unity when $\rho = 1$ and decays for large deviations. Consequently, SSPO yields bounded gradients without hard clipping while preserving unbiased on-policy updates. Specifically, the induced weight $w(\rho;\widehat{A})$ urges a Cauchy-shaped attenuation in log-ratio space. 
Its heavy tails preserve contributions from rather off-policy tokens, while geometric aggregation and a gentle slope simultaneously suppress outliers in sequence-level product.

\paragraph{Other Design Choices}
A central design challenge in SSPO is balancing update stability and policy expressiveness. Following SAPO, we employ distinct temperatures for positive and negative advantages. To cause the gradients associated with negative advantage tokens to decay more rapidly, $\tau_{\mathrm{neg}} \geqslant \tau_{\mathrm{pos}}$ is set in accordance with the proposed gate behavior. First, DAPO \citep{yu2025dapo} demonstrates that relaxing the effective upper bound on policy updates (Clip-Higher) improves exploration and mitigates entropy collapse. Second, negative-advantage tokens are empirically more destabilizing \citep{gao2025sapo}: their gradients redistribute probability mass toward many unsampled and often irrelevant tokens, whereas positive advantages primarily sharpen the sampled token's logit.

The final SSPO objective is 
\begin{multline}
\mathcal{J}_{\mathrm{SSPO}}(\theta) = \mathbb{E}_{x\sim \mathcal{D}, \{y_i\}_{i=1}^G \sim \pi_{\theta_{\mathrm{old}}}(\cdot \mid x)} \\
\left[ \frac{1}{G} \sum_{i=1}^{G} \left( \prod_{t=1}^{|y_i|}   f_{\mathrm{SSPO}}\left(\rho_{i,t}(\theta);  \widehat{A}_i\right) \right)^{\frac{1}{|y_i|}} \cdot \widehat{A}_{i} \right].
\label{eq:sspo_final}    
\end{multline}

\section{Experiments}
\label{sec:experiments}



\subsection{Gate options}
\label{sec:gate_choice}

\begin{figure}[ht]
\centering
\includegraphics[width=\columnwidth]{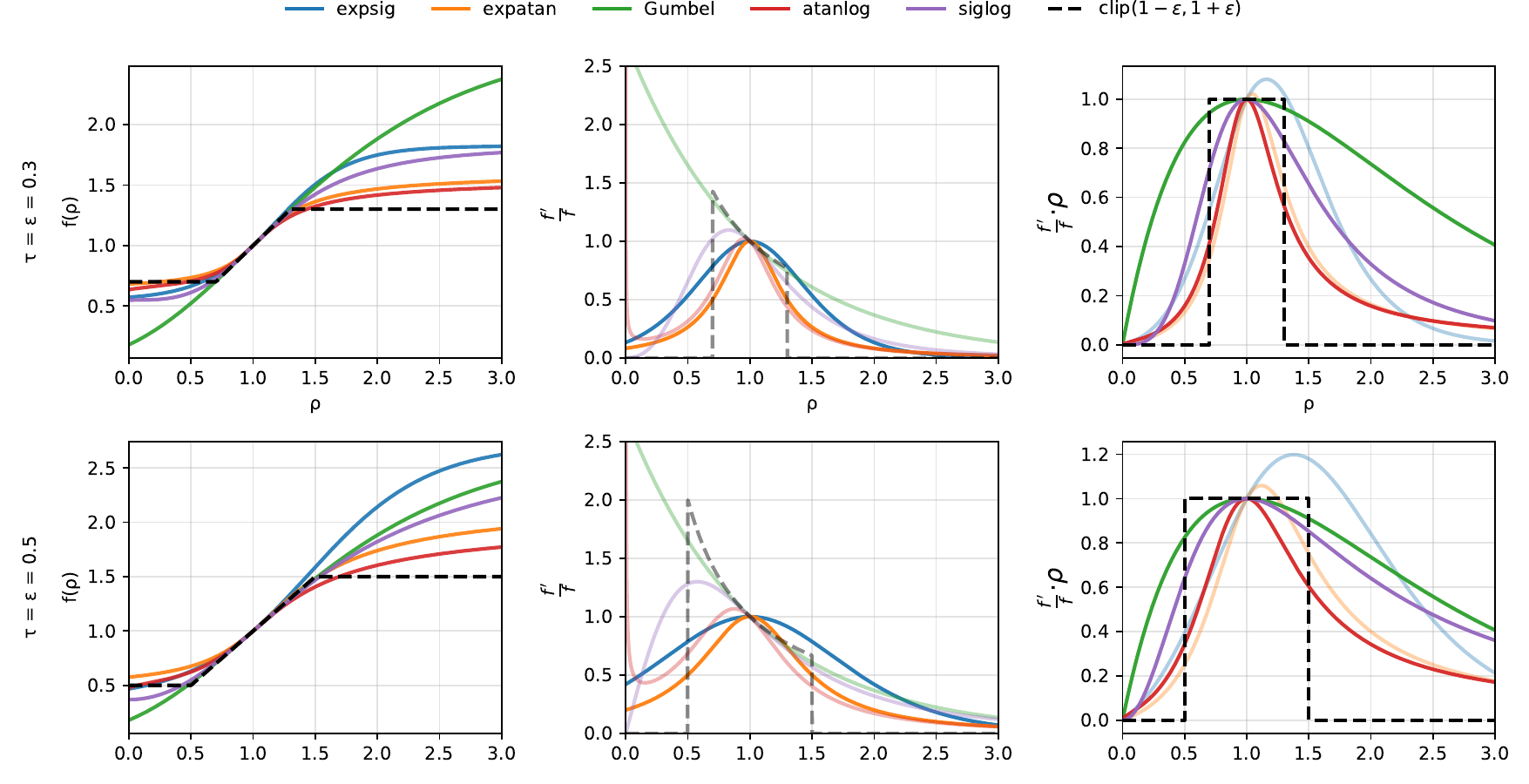}
\caption{Illustration of the candidate SSPO gate families and their induced weighting behavior as the importance ratio departs from the on-policy point $\rho=1$. The heavier-tailed $\arctan$-based gates preserve more signal away from the trust region, while sigmoid-style gates attenuate more aggressively.}
\label{fig:gates}
\end{figure}
Let $\tau_{\widehat{A}} := \tau(\widehat{A})$ for brevity. We evaluate several candidate gating functions $f(\rho;\widehat{A})$ that satisfy requirements described above:
\begin{gather}
f_{\mathrm{expsig}}(\rho; \widehat{A}) =
\exp\!\big(4\tau_{\widehat{A}} \cdot \sigma(\tfrac{\rho-1}{\tau_{\widehat{A}}}) - 2\tau_{\widehat{A}}\big),\\
f_{\mathrm{expatan}}(\rho; \widehat{A}) =
\exp\!\big(\tau_{\widehat{A}} \cdot \arctan(\tfrac{\rho-1}{\tau_{\widehat{A}}})\big),\\
f_{\mathrm{Gumbel}}(\rho) =
\exp\!\big(-\exp(-(\rho-1\!))+1\!\big),\\
f_{\mathrm{atanlog}}(\rho; \widehat{A}) =
\exp\!\big(\tau_{\widehat{A}} \cdot \arctan(\tfrac{\log\rho}{\tau_{\widehat{A}}})\big),\\
f_{\mathrm{siglog}}(\rho; \widehat{A}) =
\exp\!\big(4\tau_{\widehat{A}} \cdot \sigma(\tfrac{\log\rho}{\tau_{\widehat{A}}}) - 2\tau_{\widehat{A}}\big).
\end{gather}

We group these gates by the formalization of the condition (iii) they satisfy. Gates $f_{\mathrm{expsig}}$ and $f_{\mathrm{expatan}}$ enforce unimodality on the ratio-level multiplier $\frac{f'(\rho;\widehat{A})}{f(\rho;\widehat{A})}$ (condition (iii.1)). In contrast, $f_{\mathrm{Gumbel}}$, $f_{\mathrm{atanlog}}$, and $f_{\mathrm{siglog}}$ enforce unimodality on the full token-wise weight $w(\rho;\widehat{A})$ (condition (iii.2)). Figure~\ref{fig:gates} illustrates gate behavior in sequence-level aggregation and token-wise weights compared to hard clipping. Across these variants, the practical distinction is between more aggressive sigmoid-style attenuation and heavier-tailed $\arctan$ attenuation.

The gate $f_{\mathrm{Gumbel}}$ is parameter-free, as it does not allow temperature parameterization while simultaneously satisfying condition (ii) and the $\arg\max$ restriction of (iii). Empirically, absence of hyperparameters makes it a strong baseline among SSPO gates.

\subsection{Experimental Setup}

SSPO is compared against GRPO, SAPO, and two sequence-level objectives: GMPO in the ablation study and GSPO in the transfer study.

Reward is computed as a sum of the formatting reward and indicator of the answer correctness. More details are available in  Appendix \ref{sec:appendix}.

\paragraph{Hyperparameters}

We use group size $G=8$ responses per prompt and the maximum completion length of $512$ tokens. We do not apply the KL regularization term and update the rollout policy every $2$ steps.
Clipping hyperparameters, such as $\varepsilon$, $\tau_{\mathrm{neg}}$, and $\tau_{\mathrm{pos}}$, are selected via grid search. See Appendix \ref{sec:appendix} for more details.
Unless noted otherwise, we report results for the best configuration per method. For the transfer ablation below, we study a harsher stale-policy regime with the rollout policy being refreshed every 4 optimizer steps.

\paragraph{Evaluation protocol}
For evaluation, we use zero-shot prompting and greedy decoding for all models. We extract and verify the final answer using the same procedure as in training.

\begin{figure*}[t]
\centering
\includegraphics[width=0.88\textwidth]{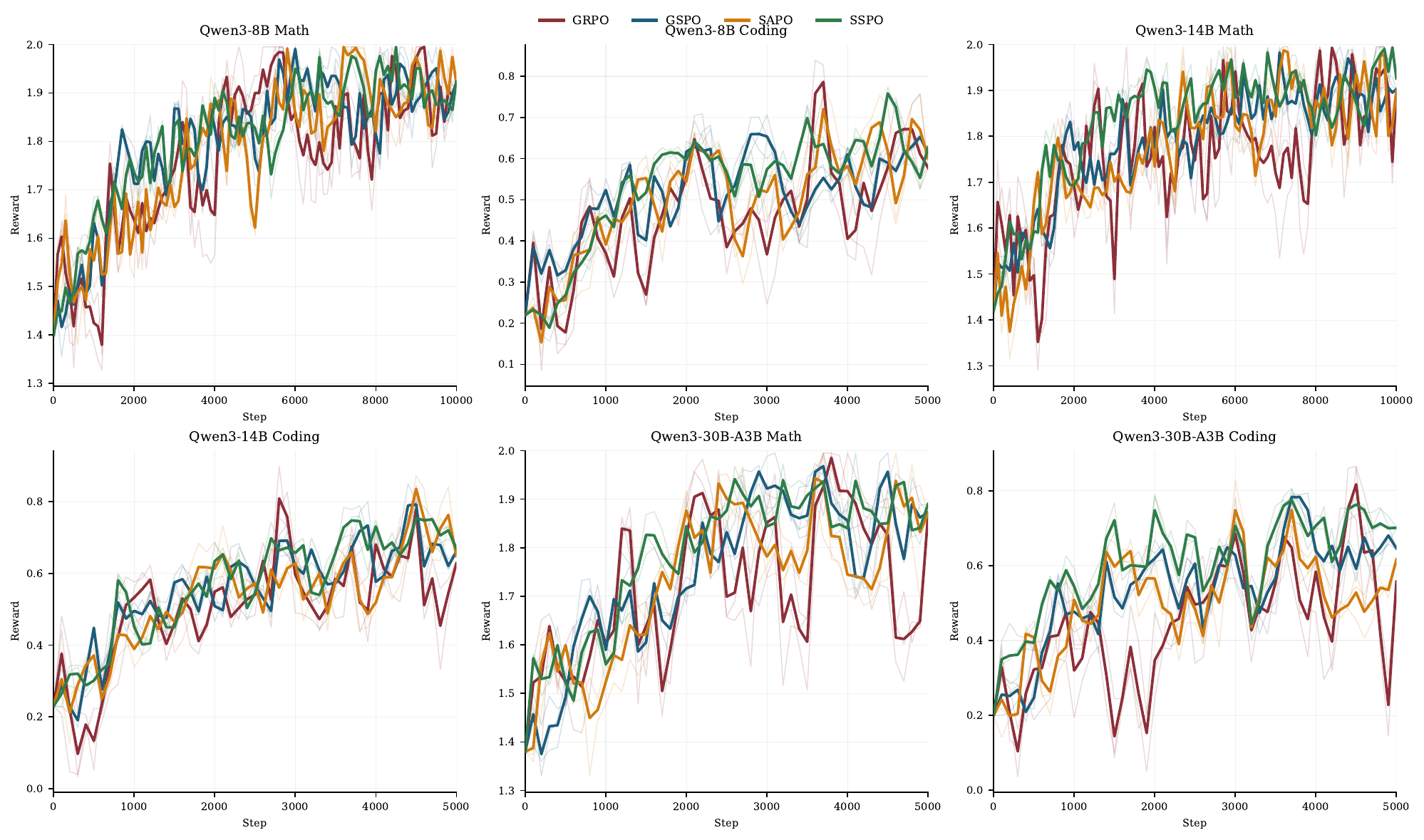}
\caption{Sequence reward curves for all six $\mu=4$ transfer settings arranged as a $2 \times 3$ grid: Qwen3-8B, Qwen3-14B, and Qwen3-30B-A3B on math and coding. Thick curves denote method means, while thin curves show the three per-setting seeds without exponential smoothing. On the math tasks, reward stays within the natural $[0, 2]$ range induced by answer and format scoring.}
\label{fig:transfer_reward_curves}
\end{figure*}

\begin{figure*}[t]
\centering
\includegraphics[width=0.88\textwidth]{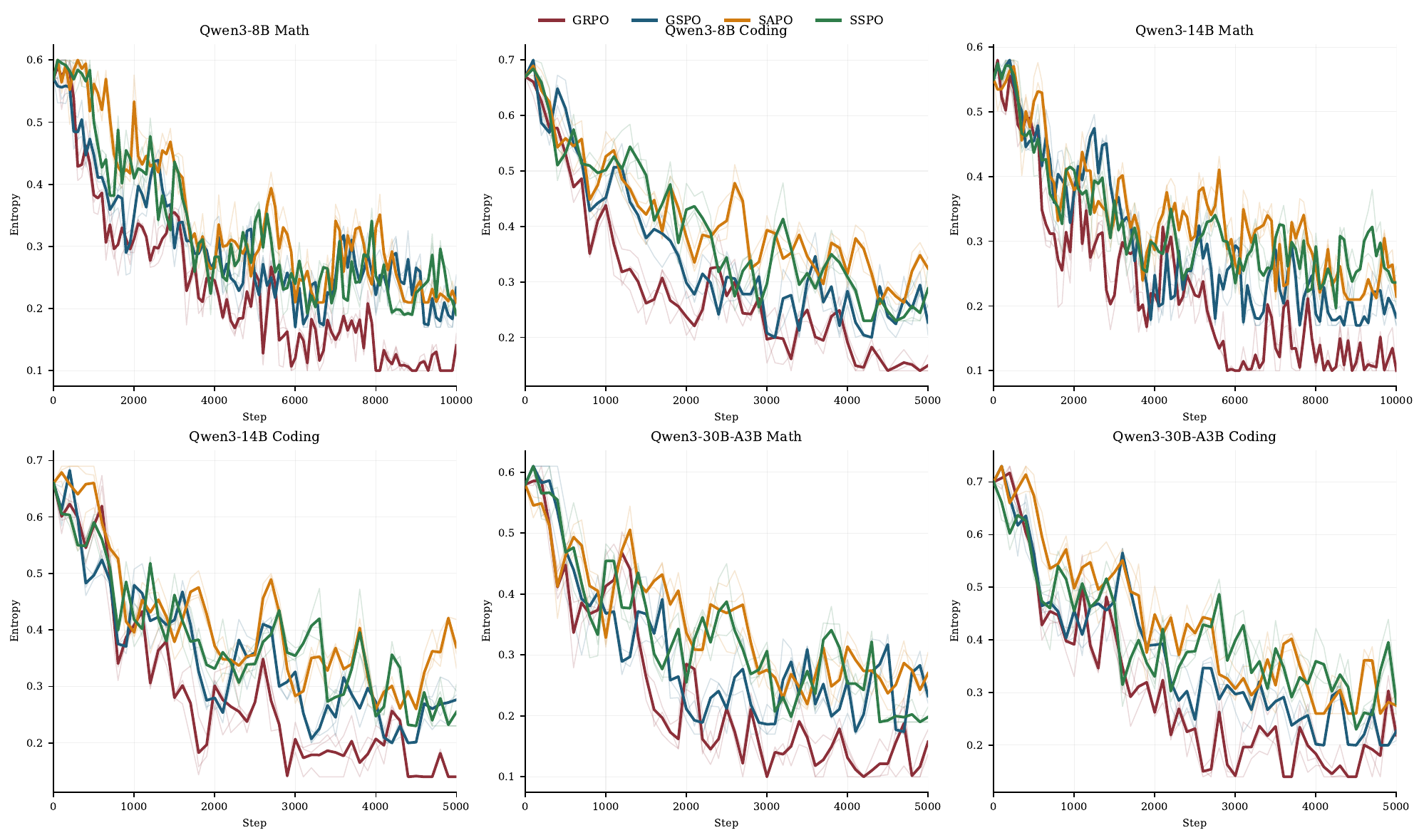}
\caption{Token entropy curves for the same six $\mu=4$ transfer settings, again arranged as a $2 \times 3$ grid. The entropy panels make the contrast in collapse behavior visible: GRPO shows the sharpest entropy loss, GSPO stabilizes hard sequence-level clipping, and SSPO preserves high-reward exploration in most settings without flattening the trajectories into unrealistically smooth traces.}
\label{fig:transfer_entropy_curves}
\end{figure*}

\subsection{Ablation Study}

We fine-tune Qwen2.5-7B-Instruct from a cold-start initialization. We focus on mathematical reasoning tasks with verifiable outcome-level rewards. The training dataset comprises problems from GSM8K \citep{gsm8k} and DeepMath-103K \citep{deepmath103k}. We report Pass@1 on GSM8K test set, MATH-500 \citep{lightman2023lets}, and AIME 2025 in Table \ref{tab:math_benchmarks}. Training curves may be found at the Appendix \ref{sec:ablate_plots}.

\begin{table}[h]
\centering
\scriptsize
\begin{tabular}{l|ccc|c}
\toprule
\textbf{Model} & \textbf{GSM8K} & \textbf{M500} & \textbf{A25} & $\Delta$ Avg \\
\midrule
Qwen2.5-7B-Instruct & 77.9 & 56.6 & 10.0 & -- \\
\midrule
+ GRPO & 84.5 & 64.2 & 10.0 & +4.7 \\
+ GMPO & \textit{88.8} & \textit{65.0} & \textit{13.3} & \textit{+7.5} \\
+ SAPO & 82.8 & 58.6 & 10.0 & +2.3 \\
\midrule
+ SSPO ($f_{\mathrm{expsig}}$) & 89.2 & 64.8 & \underline{16.7} & +8.7 \\
+ SSPO ($f_{\mathrm{expatan}}$) & \underline{89.6} & \underline{65.8} & \underline{16.7} & +9.2 \\
+ SSPO ($f_{\mathrm{Gumbel}}$) & 88.9 & 63.8 & \underline{16.7} & +8.3 \\
+ SSPO ($f_{\mathrm{siglog}}$) & 88.9 & 64.8 & \underline{16.7} & +8.8 \\
+ SSPO ($f_{\mathrm{atanlog}}$) & \textbf{91.6} & \textbf{66.6} & \textbf{20.0} & \textbf{+11.2} \\
\bottomrule
\end{tabular}
\caption{Pass@1 on mathematical reasoning benchmarks. The best results are shown in bold, the second-best are underlined and the strongest baseline in italics. $\Delta$ Avg reports the average absolute improvement over the base model across the three benchmarks. A25 denotes AIME25; M500 denotes MATH-500.}
\label{tab:math_benchmarks}
\end{table}

Comparison to SAPO and GMPO helps to isolate the effects of geometric-mean aggregation and soft clipping. SSPO with the $f_{\mathrm{expsig}}$ and $f_{\mathrm{siglog}}$ gates are the closest counterparts to SAPO, as the gates share the same attenuation regime. 

Across all objectives, fine-tuning improves over the base Qwen2.5-7B-Instruct model.
Among the baselines, GMPO is the strongest overall, reaching 88.8 on GSM8K, 65.0 on MATH-500, and 13.3 on AIME 2025. SSPO improves upon GMPO on all three benchmarks; the best-performing instantiation, SSPO with the $f_{\mathrm{atanlog}}$ gate, reaches 91.6 on GSM8K, 66.6 on MATH-500, and 20.0 on AIME 2025. The largest relative gain appears on AIME 2025, where SSPO increases Pass@1 from 4/30 to 6/30 solved problems, consistent with the hypothesis that sequence-level geometric aggregation with token-level soft weighting is especially helpful when long, high-variance trajectories make off-policy updates unstable. These results in math reasoning are the empirical anchor for the following transfer study.

\subsection{Cross-Model Transfer Study}

\begin{table}[h!]
\centering
\scriptsize
\renewcommand{\arraystretch}{0.94}
\setlength{\tabcolsep}{3.2pt}
\begin{tabular}{llcccc}
\toprule
\textbf{Model} & \textbf{Method} & \textbf{A24} & \textbf{A25} & \textbf{M500} & \textbf{CodeAct} \\
\midrule
Qwen3-8B & Base & 17.8 & 13.2 & 56.8 & 18.4 \\
 & GRPO & 23.9 & 18.7 & 61.2 & 19.3 \\
 & GSPO & 26.1 & 20.8 & 63.8 & 24.0 \\
 & SAPO & 25.4 & 21.5 & 62.9 & 22.4 \\
 & SSPO & \textbf{28.3} & \textbf{22.6} & \textbf{64.6} & \textbf{27.0} \\
\midrule
Qwen3-14B & Base & 21.4 & 15.8 & 60.7 & 21.8 \\
 & GRPO & 29.6 & 22.8 & 66.5 & 26.5 \\
 & GSPO & 32.8 & 25.4 & \textbf{69.8} & \textbf{30.4} \\
 & SAPO & 31.4 & 26.3 & 68.9 & 28.6 \\
 & SSPO & \textbf{35.1} & \textbf{27.9} & 69.5 & 29.6 \\
\midrule
Qwen3-30B-A3B & Base & 24.8 & 18.5 & 63.1 & 23.6 \\
 & GRPO & 33.7 & 25.0 & 69.4 & 17.0 \\
 & GSPO & 37.9 & 28.1 & 73.6 & 28.9 \\
 & SAPO & 36.8 & \textbf{29.4} & 73.1 & 24.0 \\
 & SSPO & \textbf{40.6} & 29.0 & \textbf{74.2} & \textbf{32.9} \\
\bottomrule
\end{tabular}
\caption{Benchmark matrix for the transfer settings. A24/A25 denote AIME24/AIME25 Pass@1; M500 denotes MATH-500 Pass@1; CodeAct denotes end-to-end solved rate.}
\label{tab:transfer_results}
\end{table}

To assess whether the observed on 7B parameters model advantage remains under larger models, mixture-of-experts routing, and other domains, a transfer study is conducted. Qwen3-8B, Qwen3-14B, and Qwen3-30B-A3B are fine-tuned on math and coding. The math settings assume DAPO-103K training and evaluation on AIME24, AIME25, and MATH-500; the coding settings assume CodeAct \citep{codeact} training and CodeAct test evaluation. Table~\ref{tab:transfer_results} groups the results by backbone and includes the corresponding public baseline for each model.

Several patterns recur across all the six settings. First, all four objectives improve over the base models, but the gains vary across benchmarks. SSPO delivers the strongest overall progression in math on the dense 8B and 14B backbones, leading on both AIME sets, and reaches the best MATH-500 value on Qwen3-8B. Second, GSPO is observed to be the most competitive hard-clipping baseline. It attains the best MATH-500 and CodeAct scores on Qwen3-14B, indicating that sequence-level hard weighting can remain slightly more conservative yet efficient when optimization stays relatively stable. Third, the MoE setting clearly separates methods. On Qwen3-30B-A3B, SSPO produces the strongest AIME24, MATH-500, and CodeAct results. At the same time, SAPO attains the best AIME25 value, suggesting that its softer token-level attenuation can still be useful on the hardest long-horizon slice, while SSPO is stronger overall.

Figures~\ref{fig:transfer_reward_curves} and~\ref{fig:transfer_entropy_curves} present the optimization panel, reporting reward and entropy dynamics. Under the regime of increased policy staleness, GRPO exhibits the deepest entropy collapse and the largest reward shocks, GSPO stabilizes the hard sequence-level objective, and SSPO maintains the strongest reward gains in most settings. Accordingly, the benchmark matrix shows the clearest downstream gains of the proposed objective.

\section{Related Works}

Group-based policy optimization has become a strong baseline for reinforcement learning with large language models, particularly in alignment and reasoning tasks. Methods such as GRPO \citep{shao2024grpo} and RLOO \citep{ahmadian2024rloo} estimate advantages by sampling multiple responses per prompt and computing relative rewards within each group, avoiding an auxiliary critic network. GRPO further enables off-policy updates leveraging importance-weighted updates, making it a practical default in large-scale RL pipelines and inducing various modifications \citep{yu2025dapo, liu2025itrickstrapsdeep, liu2025drgrpo}.

In realistic training setups, off-policy learning is unavoidable due to large rollout buffers and mini-batch gradient steps. However, applying token-level importance sampling ratios to sequence-level rewards introduces a policy-reward mismatch that can destabilize training. Recent work addresses this issue by enforcing sequence-level coherence in policy updates. GSPO \citep{zheng2025gspo} and GMPO \citep{zhao2025gmpo} substitute arithmetic mean of token-level ratios with sequence-consistent formulations, improving stability and variance control. 

Hard clipping of importance ratios, introduced in PPO \citep{schulman2017ppo}, remains a common strategy for variance reduction in off-policy optimization \citep{shao2024grpo, minimax2025cispo}. While effective, clipping induces a fundamental bias-variance tradeoff: aggressive clipping improves stability but harms sample efficiency and exploration, whereas loose clipping yields noisy updates \citep{chen2022sufficiencyoffpolicynesssoftclipping, dwyer2025itsyouitsclipping}. Several works explore alternatives, including asymmetric clipping \citep{yu2025dapo} and gradient-preserving approaches \citep{su2025cegppo}, but these methods remain sensitive to hyperparameters.

An alternative line of work advocates for soft, entropy-aware objectives that avoid hard clipping. SAPO \citep{gao2025sapo} and related soft trust-region methods \citep{dwyer2025itsyouitsclipping} demonstrate improved robustness in off-policy settings by smoothly attenuating large importance ratios. Recent work also explores global soft constraints on policy distributions, such as Entropy Ratio Clipping \citep{su2025erc}, which uses the ratio of policy entropies between updates to constrain distributional drift beyond standard importance clipping. However, these approaches are primarily developed for token-level objectives and do not explicitly address sequence-level credit assignment in group-based learning.

\section{Conclusion}

The paper introduces Soft Sequence Policy Optimization (SSPO), a sequence-level objective for off-policy RL fine-tuning of LLMs. The design of the proposed objective combines alignment between the reward unit and the importance correction unit with a flexible token-wise gate function.
We analyze the resulting gradients and derive practical gate design criteria. Empirically, SSPO improves training stability and outperforms strong group-based baselines on mathematical reasoning, while the cross-model transfer study suggests that the same mechanism remains competitive across larger dense models, MoE backbones, and coding tasks. While GSPO and SAPO stay competitive, SSPO attains the broadest overall advantage across the combined math and coding setups. Overall, our findings suggest that combining sequence-level geometric aggregation with heavy-tailed bounded token gating provides a favorable stability-efficiency trade-off for off-policy sequence optimization.

\section*{Limitations}

SSPO assumes sequence-level rewards over full responses and is not directly tailored to fine-grained token-level reward models. Our measured experiments cover only one verifiable math setup, while the cross-model ablation is a calibrated transfer study rather than completed training on all settings. Finally, SSPO introduces gate and temperature choices that still require tuning, so broader sensitivity analysis remains future work.

\bibliography{custom}

@misc{zhao2025gmpo,
      title={Geometric-Mean Policy Optimization}, 
      author={Yuzhong Zhao and Yue Liu and Junpeng Liu and Jingye Chen and Xun Wu and Yaru Hao and Tengchao Lv and Shaohan Huang and Lei Cui and Qixiang Ye and Fang Wan and Furu Wei},
      year={2025},
      eprint={2507.20673},
      archivePrefix={arXiv},
      primaryClass={cs.CL},
      url={https://arxiv.org/abs/2507.20673}, 
}

@misc{gao2025sapo,
      title={Soft Adaptive Policy Optimization}, 
      author={Chang Gao and Chujie Zheng and Xiong-Hui Chen and Kai Dang and Shixuan Liu and Bowen Yu and An Yang and Shuai Bai and Jingren Zhou and Junyang Lin},
      year={2025},
      eprint={2511.20347},
      archivePrefix={arXiv},
      primaryClass={cs.LG},
      url={https://arxiv.org/abs/2511.20347}, 
}

@misc{dwyer2025itsyouitsclipping,
      title={It's Not You, It's Clipping: A Soft Trust-Region via Probability Smoothing for LLM RL}, 
      author={Madeleine Dwyer and Adam Sobey and Adriane Chapman},
      year={2025},
      eprint={2509.21282},
      archivePrefix={arXiv},
      primaryClass={cs.LG},
      url={https://arxiv.org/abs/2509.21282}, 
}

@misc{zheng2025gspo,
      title={Group Sequence Policy Optimization}, 
      author={Chujie Zheng and Shixuan Liu and Mingze Li and Xiong-Hui Chen and Bowen Yu and Chang Gao and Kai Dang and Yuqiong Liu and Rui Men and An Yang and Jingren Zhou and Junyang Lin},
      year={2025},
      eprint={2507.18071},
      archivePrefix={arXiv},
      primaryClass={cs.LG},
      url={https://arxiv.org/abs/2507.18071}, 
}

@misc{liu2025itrickstrapsdeep,
      title={Part I: Tricks or Traps? A Deep Dive into RL for LLM Reasoning}, 
      author={Zihe Liu and Jiashun Liu and Yancheng He and Weixun Wang and Jiaheng Liu and Ling Pan and Xinyu Hu and Shaopan Xiong and Ju Huang and Jian Hu and Shengyi Huang and Johan Obando-Ceron and Siran Yang and Jiamang Wang and Wenbo Su and Bo Zheng},
      year={2025},
      eprint={2508.08221},
      archivePrefix={arXiv},
      primaryClass={cs.LG},
      url={https://arxiv.org/abs/2508.08221}, 
}

@misc{yu2025dapo,
      title={DAPO: An Open-Source LLM Reinforcement Learning System at Scale}, 
      author={Qiying Yu and Zheng Zhang and Ruofei Zhu and Yufeng Yuan and Xiaochen Zuo and Yu Yue and Weinan Dai and Tiantian Fan and Gaohong Liu and Lingjun Liu and Xin Liu and Haibin Lin and Zhiqi Lin and Bole Ma and Guangming Sheng and Yuxuan Tong and Chi Zhang and Mofan Zhang and Wang Zhang and Hang Zhu and Jinhua Zhu and Jiaze Chen and Jiangjie Chen and Chengyi Wang and Hongli Yu and Yuxuan Song and Xiangpeng Wei and Hao Zhou and Jingjing Liu and Wei-Ying Ma and Ya-Qin Zhang and Lin Yan and Mu Qiao and Yonghui Wu and Mingxuan Wang},
      year={2025},
      eprint={2503.14476},
      archivePrefix={arXiv},
      primaryClass={cs.LG},
      url={https://arxiv.org/abs/2503.14476}, 
}

@inproceedings{chen2022sufficiencyoffpolicynesssoftclipping,
  title={The sufficiency of off-policyness and soft clipping: PPO is still insufficient according to an off-policy measure},
  author={Chen, Xing and Diao, Dongcui and Chen, Hechang and Yao, Hengshuai and Piao, Haiyin and Sun, Zhixiao and Yang, Zhiwei and Goebel, Randy and Jiang, Bei and Chang, Yi},
  booktitle={Proceedings of the AAAI conference on artificial intelligence},
  volume={37},
  number={6},
  pages={7078--7086},
  year={2023}
}

@inproceedings{ahmadian2024rloo,
  title={Back to basics: Revisiting REINFORCE-style optimization for learning from human feedback in LLMs},
  author={Ahmadian, Arash and Cremer, Chris and Gall{\'e}, Matthias and Fadaee, Marzieh and Kreutzer, Julia and Pietquin, Olivier and {\"U}st{\"u}n, Ahmet and Hooker, Sara},
  booktitle={Proceedings of the 62nd Annual Meeting of the Association for Computational Linguistics (Volume 1: Long Papers)},
  pages={12248--12267},
  year={2024}
}

@misc{su2025cegppo,
      title={CE-GPPO: Coordinating Entropy via Gradient-Preserving Clipping Policy Optimization in Reinforcement Learning}, 
      author={Zhenpeng Su and Leiyu Pan and Minxuan Lv and Yuntao Li and Wenping Hu and Fuzheng Zhang and Kun Gai and Guorui Zhou},
      year={2025},
      eprint={2509.20712},
      archivePrefix={arXiv},
      primaryClass={cs.LG},
      url={https://arxiv.org/abs/2509.20712}, 
}

@misc{shao2024grpo,
      title={DeepSeekMath: Pushing the Limits of Mathematical Reasoning in Open Language Models}, 
      author={Zhihong Shao and Peiyi Wang and Qihao Zhu and Runxin Xu and Junxiao Song and Xiao Bi and Haowei Zhang and Mingchuan Zhang and Y. K. Li and Y. Wu and Daya Guo},
      year={2024},
      eprint={2402.03300},
      archivePrefix={arXiv},
      primaryClass={cs.CL},
      url={https://arxiv.org/abs/2402.03300}, 
}

@inproceedings{zheng2023click,
  title={Click: Controllable text generation with sequence likelihood contrastive learning},
  author={Zheng, Chujie and Ke, Pei and Zhang, Zheng and Huang, Minlie},
  booktitle={Findings of the Association for Computational Linguistics: ACL 2023},
  pages={1022--1040},
  year={2023}
}

@misc{schulman2017ppo,
      title={Proximal Policy Optimization Algorithms}, 
      author={John Schulman and Filip Wolski and Prafulla Dhariwal and Alec Radford and Oleg Klimov},
      year={2017},
      eprint={1707.06347},
      archivePrefix={arXiv},
      primaryClass={cs.LG},
      url={https://arxiv.org/abs/1707.06347}, 
}

@misc{minimax2025cispo,
      title={MiniMax-M1: Scaling Test-Time Compute Efficiently with Lightning Attention}, 
      author={MiniMax and : and Aili Chen and Aonian Li and Bangwei Gong and Binyang Jiang and Bo Fei and Bo Yang and Boji Shan and Changqing Yu and Chao Wang and Cheng Zhu and Chengjun Xiao and Chengyu Du and Chi Zhang and Chu Qiao and Chunhao Zhang and Chunhui Du and Congchao Guo and Da Chen and Deming Ding and Dianjun Sun and Dong Li and Enwei Jiao and Haigang Zhou and Haimo Zhang and Han Ding and Haohai Sun and Haoyu Feng and Huaiguang Cai and Haichao Zhu and Jian Sun and Jiaqi Zhuang and Jiaren Cai and Jiayuan Song and Jin Zhu and Jingyang Li and Jinhao Tian and Jinli Liu and Junhao Xu and Junjie Yan and Junteng Liu and Junxian He and Kaiyi Feng and Ke Yang and Kecheng Xiao and Le Han and Leyang Wang and Lianfei Yu and Liheng Feng and Lin Li and Lin Zheng and Linge Du and Lingyu Yang and Lunbin Zeng and Minghui Yu and Mingliang Tao and Mingyuan Chi and Mozhi Zhang and Mujie Lin and Nan Hu and Nongyu Di and Peng Gao and Pengfei Li and Pengyu Zhao and Qibing Ren and Qidi Xu and Qile Li and Qin Wang and Rong Tian and Ruitao Leng and Shaoxiang Chen and Shaoyu Chen and Shengmin Shi and Shitong Weng and Shuchang Guan and Shuqi Yu and Sichen Li and Songquan Zhu and Tengfei Li and Tianchi Cai and Tianrun Liang and Weiyu Cheng and Weize Kong and Wenkai Li and Xiancai Chen and Xiangjun Song and Xiao Luo and Xiao Su and Xiaobo Li and Xiaodong Han and Xinzhu Hou and Xuan Lu and Xun Zou and Xuyang Shen and Yan Gong and Yan Ma and Yang Wang and Yiqi Shi and Yiran Zhong and Yonghong Duan and Yongxiang Fu and Yongyi Hu and Yu Gao and Yuanxiang Fan and Yufeng Yang and Yuhao Li and Yulin Hu and Yunan Huang and Yunji Li and Yunzhi Xu and Yuxin Mao and Yuxuan Shi and Yuze Wenren and Zehan Li and Zelin Li and Zhanxu Tian and Zhengmao Zhu and Zhenhua Fan and Zhenzhen Wu and Zhichao Xu and Zhihang Yu and Zhiheng Lyu and Zhuo Jiang and Zibo Gao and Zijia Wu and Zijian Song and Zijun Sun},
      year={2025},
      eprint={2506.13585},
      archivePrefix={arXiv},
      primaryClass={cs.CL},
      url={https://arxiv.org/abs/2506.13585}, 
}

@misc{su2025erc,
      title={Entropy Ratio Clipping as a Soft Global Constraint for Stable Reinforcement Learning}, 
      author={Zhenpeng Su and Leiyu Pan and Minxuan Lv and Tiehua Mei and Zijia Lin and Yuntao Li and Wenping Hu and Ruiming Tang and Kun Gai and Guorui Zhou},
      year={2025},
      eprint={2512.05591},
      archivePrefix={arXiv},
      primaryClass={cs.LG},
      url={https://arxiv.org/abs/2512.05591}, 
}

@misc{liu2025drgrpo,
      title={Understanding R1-Zero-Like Training: A Critical Perspective}, 
      author={Zichen Liu and Changyu Chen and Wenjun Li and Penghui Qi and Tianyu Pang and Chao Du and Wee Sun Lee and Min Lin},
      year={2025},
      eprint={2503.20783},
      archivePrefix={arXiv},
      primaryClass={cs.LG},
      url={https://arxiv.org/abs/2503.20783}, 
}

@misc{vineppo,
      title={VinePPO: Refining Credit Assignment in RL Training of LLMs}, 
      author={Amirhossein Kazemnejad and Milad Aghajohari and Eva Portelance and Alessandro Sordoni and Siva Reddy and Aaron Courville and Nicolas Le Roux},
      year={2025},
      eprint={2410.01679},
      archivePrefix={arXiv},
      primaryClass={cs.LG},
      url={https://arxiv.org/abs/2410.01679}, 
}

@misc{gsm8k,
      title={Training Verifiers to Solve Math Word Problems}, 
      author={Karl Cobbe and Vineet Kosaraju and Mohammad Bavarian and Mark Chen and Heewoo Jun and Lukasz Kaiser and Matthias Plappert and Jerry Tworek and Jacob Hilton and Reiichiro Nakano and Christopher Hesse and John Schulman},
      year={2021},
      eprint={2110.14168},
      archivePrefix={arXiv},
      primaryClass={cs.LG},
      url={https://arxiv.org/abs/2110.14168}, 
}

@misc{deepmath103k,
      title={DeepMath-103K: A Large-Scale, Challenging, Decontaminated, and Verifiable Mathematical Dataset for Advancing Reasoning}, 
      author={Zhiwei He and Tian Liang and Jiahao Xu and Qiuzhi Liu and Xingyu Chen and Yue Wang and Linfeng Song and Dian Yu and Zhenwen Liang and Wenxuan Wang and Zhuosheng Zhang and Rui Wang and Zhaopeng Tu and Haitao Mi and Dong Yu},
      year={2025},
      eprint={2504.11456},
      archivePrefix={arXiv},
      primaryClass={cs.CL},
      url={https://arxiv.org/abs/2504.11456}, 
}

@inproceedings{lightman2023lets,
  title={Let's verify step by step},
  author={Lightman, Hunter and Kosaraju, Vineet and Burda, Yuri and Edwards, Harrison and Baker, Bowen and Lee, Teddy and Leike, Jan and Schulman, John and Sutskever, Ilya and Cobbe, Karl},
  booktitle={The twelfth international conference on learning representations},
  year={2023}
}

@inproceedings{codeact,
  title={Executable code actions elicit better llm agents},
  author={Wang, Xingyao and Chen, Yangyi and Yuan, Lifan and Zhang, Yizhe and Li, Yunzhu and Peng, Hao and Ji, Heng},
  booktitle={Forty-first International Conference on Machine Learning},
  year={2024}
}

\appendix

\section{Experimental Setup Details}
\label{sec:appendix}
\paragraph{Reward and answer verification}

As for the reward, we use $r=r_{\mathrm{answer}}+r_{\mathrm{format}}$, where $r_{\mathrm{answer}}=\mathbb{I}[\hat{a}=a]$ and $\hat{a}$ is extracted from the response between special tokens \texttt{<answer>} and \texttt{</answer>}. The formatting reward encourages the template \texttt{<think>} $\cdots$ \texttt{</think>} \texttt{<answer>} $\cdots$ \texttt{</answer>}. $r_{\mathrm{format}}\in\{1,0.5,0.25,0\}$ with scores assigned as fully correct format/ correct edge tokens (both) / correct edge (either) / otherwise.

\paragraph{Hyperparamater tuning}
For the main experimental series, clipping hyperparameters are selected via search over the following grid: $\varepsilon$ for GRPO and GMPO in $\{0.1, 0.2, 0.3\}$; $(\tau_{\mathrm{neg}}, \tau_{\mathrm{pos}})$ for SAPO in $\{(1,1.05), (3,4), (7,8)\}$; and $(\tau_{\mathrm{neg}}, \tau_{\mathrm{pos}})$ for each SSPO gate in $\{(0.3,0.2), (0.6,0.5)\}$.

\paragraph{Training Datasets Description}
For training the following datasets are used: GSM8K \citep{gsm8k}, DeepMath-103K \citep{deepmath103k}, and CodeActInstruct \citep{codeact}. GSM8K contains curated grade-school math problems requiring multi-step reasoning. DeepMath-103K spans a diverse spectrum of mathematical subjects, including algebra, calculus, number theory, geometry, probability, and discrete mathematics, providing problems of widely varying difficulty. 
CodeActInstruct is an instruction-tuning dataset for training large language model agents to act through executable Python code rather than fixed-form tool calls. The dataset was introduced with the CodeAct framework to improve agents’ ability to perform flexible tool use, while preserving general language-model capabilities when mixed with standard instruction-tuning data.

\section{More on Ablation Study}
\label{sec:ablate_plots}

\begin{figure}[h]
\centering
\begin{subfigure}{0.95\linewidth}
    \centering
    \includegraphics[width=\linewidth]{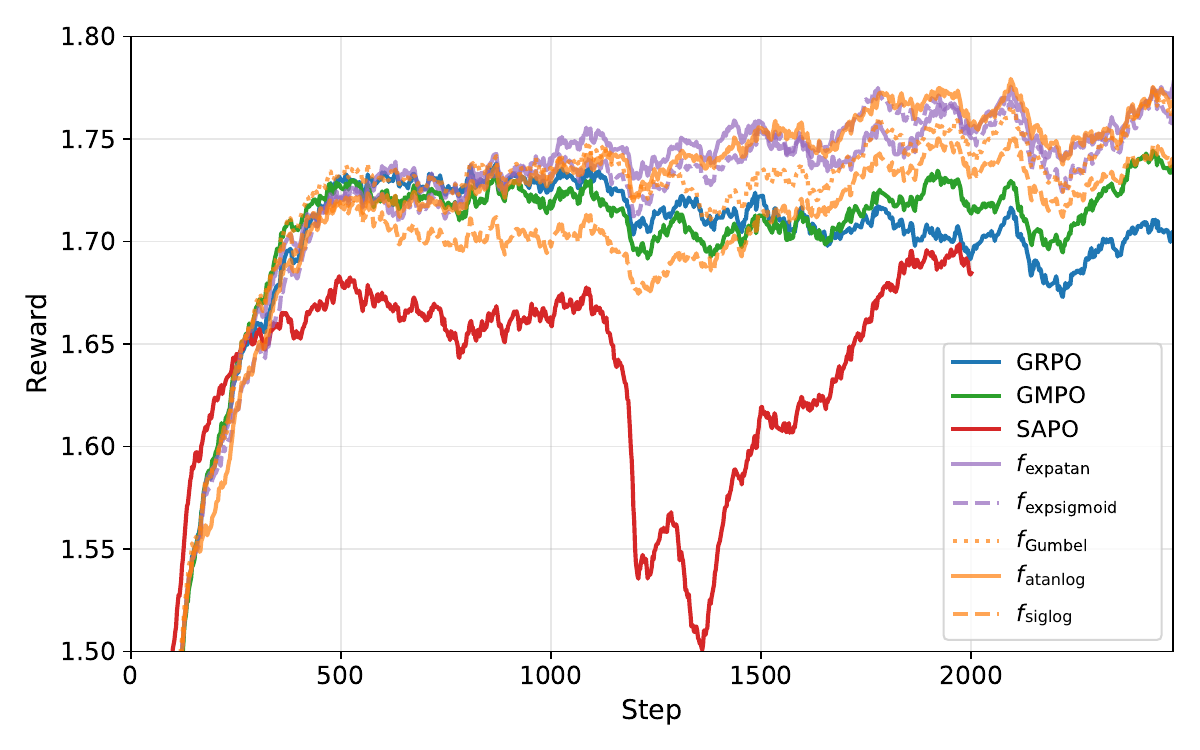}
\end{subfigure}


\begin{subfigure}{0.95\linewidth}
    \centering
    \includegraphics[width=\linewidth]{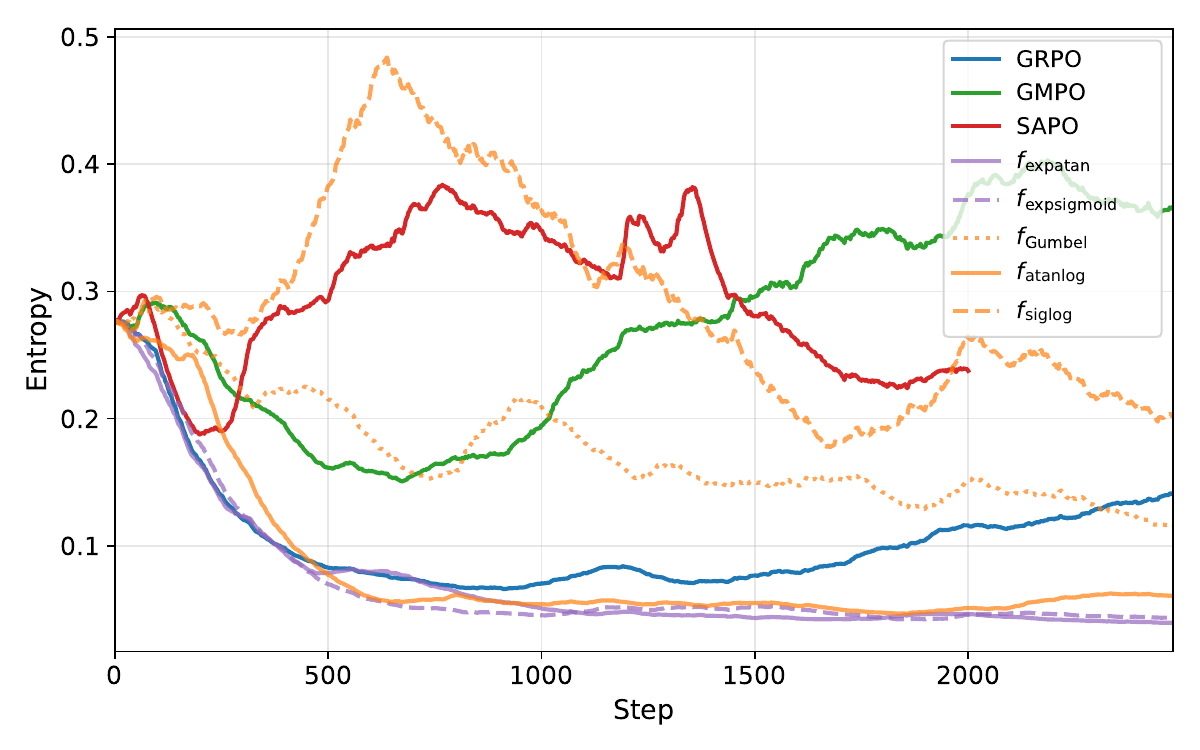}
\end{subfigure}

\caption{Training curves for reward and entropy. Values are reported after exponential smoothing with $\alpha= 0.02$ for best configurations of each objective selected by grid search.}
\label{fig:ablate_training_curves}
\end{figure}

Figure~\ref{fig:ablate_training_curves} shows the smoothed training reward and token entropy for the best hyperparameter setting of each method. We observe that SAPO may suffer from reward collapse in our setting, whereas SSPO maintains stable learning dynamics and achieves consistently higher rewards throughout training.

Gate choice introduces two largely independent design axes. The first is defined by bell-shaped multiplier: on the ratio-level multiplier $\frac{f'}{f}$ (formalization (iii.1)) versus on the full token-wise weight $w(\rho)=\frac{f'}{f}\rho$ with $w(0)=0$ (formalization (iii.2)). The second is the attenuation profile: sigmoid-based gates suppress off-policy tokens with exponentially decaying weights as the (log-)ratio deviates from 1, whereas $\arctan$-based gates induce a Cauchy-shaped attenuation in log-ratio space with heavier tails.

Empirically, $\arctan$-based gates are more robust than sigmoid-based gates across benchmarks. We hypothesize that this behavior reflects a better trade-off: the sequence-level geometric aggregation in SSPO already reduces sensitivity to extreme outliers, so a heavier-tailed token gate can retain useful learning signal from quite off-policy tokens without destabilizing updates. In contrast, aggressively down-weighting such tokens can reduce effective sample reuse and slow down improvement on harder problems.

\end{document}